# System-level Impact of Non-Ideal Program-Time of Charge Trap Flash (CTF) on Deep Neural Network

S. Shrivastava, A. Biswas, S. Chakrabarty, G. Dash, V. Saraswat, and U. Ganguly, *Member, IEEE*

**Abstract**—Learning of deep neural networks (DNN) using Resistive Processing Unit (RPU) architecture is energy-efficient as it utilizes dedicated neuromorphic hardware and stochastic computation of weight updates for in-memory computing. Charge Trap Flash (CTF) devices can implement RPU-based weight updates in DNNs. However, prior work has shown that the weight updates ($V_T$) in CTF-based RPU are impacted by the non-ideal program time of CTF. The non-ideal program time is affected by two factors of CTF. Firstly, the effects of the number of input pulses (N) or pulse width (pw), and secondly, the gap between successive update pulses ($t_{gap}$) used for the stochastic computation of weight updates. Therefore, the impact of this non-ideal program time must be studied for neural network training simulations. In this study, Firstly, we propose a pulse-train design compensation technique to reduce the total error caused by non-ideal program time of CTF and stochastic variance of a network. Secondly, we simulate RPU-based DNN with non-ideal program time of CTF on MNIST and Fashion-MNIST datasets. We find that for larger N (~1000), learning performance approaches the ideal (software-level) training level and, therefore, is not much impacted by the choice of $t_{gap}$ used to implement RPU-based weight updates. However, for lower N (<500), learning performance depends on $T_{gap}$ of the pulses. Finally, we also performed an ablation study to isolate the causal factor of the improved learning performance. We conclude that the lower noise level in the weight updates is the most likely significant factor to improve the learning performance of DNN. Thus, our study attempts to compensate for the error caused by non-ideal program time and standardize the pulse length (N) and pulse gap ($t_{gap}$) specifications for CTF-based RPUs for accurate system-level on-chip training.

This work is supported in parts by the Department of Science and Technology (DST) Nano Mission, Ministry of Electronics, and IT (MeitY) and Department of Electronics through the Nanoelectronics Network for Research and Applications (NNETRA) project, Govt. of India. Vivek Saraswat is supported by Prime Minister's Research Fellowship, Govt. of India.

Shalini Shrivastava (shalinishri.iit@gmail.com), Anmol Biswas (anmolbiswas@gmail.com), Gayatri Dash (gdash1110@gmail.com), Vivek Saraswat (vsaraswat009@gmail.com) and Udayan Ganguly (udayan@ee.iitb.ac.in) are with Department of Electrical Engineering, Indian Institute of Technology Bombay, Mumbai - 400076, Maharashtra, India. Samyak Chakrabarty (skcy2001@gmail.com) is with Department of Electronics and Electrical Communications Engineering, Indian Institute of Technology Kharagpur, Kharagpur – 721302, West Bengal, India. Udayan Ganguly is also with the Centre for Semiconductor Technologies (SemiX), IIT Bombay, Mumbai – 400076, Maharashtra, India

## I. Introduction

ARTIFICIAL Intelligence has exceptional skills in translation, recognition, perception, and solving many cognitive tasks by using Deep Learning [1]. Deep learning is performed by training a deep neural network (DNN) with multiple hidden layers between the input and the output layers. These networks are trained by supervised learning algorithms like back-propagation to enable gradient descent [2]. Back-propagation requires the inner product of activation ($x_i$) of the $i$th neuron of the input layer and error ($\delta_j$) of the $j$th neuron of the output layer to compute the weight update i.e. $\Delta w_{ij} \propto x_i.\delta_j$ [3]. For $n$ neurons in both the input layer and output layers, the weight update is an $O(n^2)$ multiplication operation [4]. Thus, the training of DNNs is a time-consuming as well as an energy-intensive process [5]. Resistive Processing Unit (RPU) architecture has been proposed for energy-efficient training of DNNs [6]. It consists of two major units:

1) Stochastic Pulse Generation: It encodes "normalized" $x_i$ and $\delta_j$ as probabilities of identical pulses in a $N$-slot pulse train where $x_i, \delta_j \in$ (0,1) (Fig. 1a). The $P_{x_i}$ and $P_{\delta_j}$ are stochastic input programming pulses. The overlapped of these input pulses due to AND operation results in the net input pulse. The $N$-slot net input pulse train will produce $\sim N.x_i.\delta_j$ program pulses for weight-update by analog memory element of the RPU architecture [6].

2) Analog Memory Unit (AMU): It performs multiplication by AND operation, and thresholding, followed by weight-update in the RPU architecture. For correct weight update, the AMU should have the following requirements: (i) The weight update must have excellent linearity with pulse number i.e. $\Delta w_{ij} \propto N.x_i.\delta_j$; (ii) Gradual weight update enables larger $N$ stochastic weight updates, and thus it reduces stochastic variance $\Delta_{stoch} \propto 1/\sqrt{N}$ based on Poisson statistics. (iii) Nonlinear thresholding for programming so that the weight update should happen only for overlapped input pulses ($P_{x_i}.P_{\delta_j}$) and (iv), the memory should have a large conductance range ($g_{max} - g_{min}$) compared to the variations/noise in weight update (Signal to Noise) [6]. Recently the Charge Trap Flash (CTF) has been demonstrated as an excellent choice for RPU elements with all the above properties [7].

In RPU architecture N varies in order to reduce the stochastic gradient. This division of an input pulse to an N-slot pulse train is a common approach for many in-memory computing training architectures like the Resistive Processing Unit (RPU). In our earlier work, we assumed that the weight change, i.e., $V_T$ is conserved with the division of N. The $V_T$ change of CTF due to pulsing has been modeled as a weight update and has been used to predict software-level accuracy using RPU-based learning [7]. However, recently we have demonstrated that $V_T$ is unconserved with the division of N [8]. This is due to the accumulation of charges in traps of blocking oxide of CTF when the trains of different lengths pulses ($N$) are applied. This is significant in determining the final weight update. In general, when the total ON-time, $T_{ON}$, is divided into $N$ discrete program pulses, the resulting pulse train has pulse width (pw) and the gap between the pulses as $T_{gap}$. In [8], we have demonstrated that due to charge accumulation in blocking oxide (BO) of CTF, $V_T$ is not conserved with $N$(or $t_{pw}$), and $t_{gap}$, resulting in non-ideal program time Fig 1©. This also impacts $V_T$ and stochastic variance $\Delta_{stoch} \propto 1/\sqrt{N}$ based on Poisson statistics. The non-ideal program time impacts the weight-update, and thus it may affect the system-level performance of the network in RPU architecture.

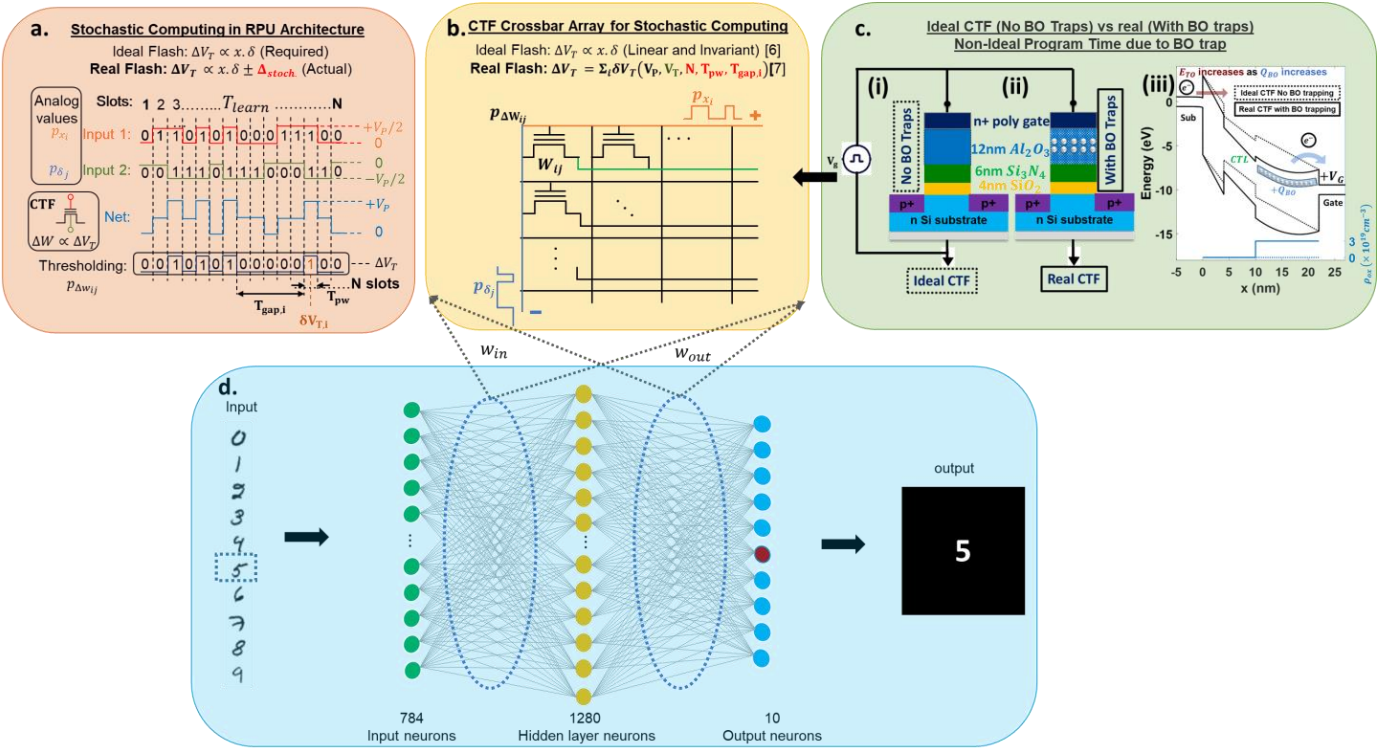

Fig. 1. CTF in RPU architecture for Neural network training: (a) Stochastic computing in RPU architecture using CTF as an RPU element (b) CTF crossbar array for stochastic computing, explaining that the ideal flash is linear and invariant whereas in real flash weight update is the function of input pulses (N as discrete number of pulses, pulse width (pw) and gap between the pulses (t_gap). (c) (i) Ideal (No BO traps) vs (ii) Non-Ideal (with BO traps) CTF, iii. The difference in the fields of the band diagram of CTF is due to BO traps depend on the input pulses Tgap and N (d) Deep neural network (DNN) with 784 input, 10 output, and 1280 hidden layers neurons. The weight updates in the network are stochastic multiplication computed by a crossbar array.

To accurately predict the system-level performance of learning algorithms running with CTF-based RPU architecture, it is essential to understand the impact of non-ideal program time on *a)* the stochastic variance of a CTF device and compensate for it, and *(b)* the performance of the neural network learning consisting of CTF-based devices. Hence, we first study the impact of non-ideal program time on the stochastic variance of a CTF device. Secondly, we characterize and study the experimental non-ideal program time of a CTF device on deep learning for neural networks.

Thus, in this paper, firstly, we analyze the total error due to stochastic encoding error and non-ideal program time error and then compensate for it by proposing different pulse schemes. Secondly, we implemented the sophisticated models of the CTF non-ideal program time and finally, we analyzed their impact on training algorithms using RPU architecture for MNIST [9] & and FMNIST [10] datasets. Hence our study enhances the system-level feasibility of in-memory functional computation for hardware acceleration using CTF-based RPU architecture for large-scale DNNs.

## II. CHARGE TRAP FLASH AND ITS NON-IDEAL PROGRAM TIME

To analyze the effect of non-ideal program time of CTF on weight updates in RPU-based neural network training, we use the CTF device as described in [8]. This Silicon Nitride-based CTF has an n-Si substrate and Boron-doped p+ source/drain regions. The gate stack consists of an- $SiO_2$ tunnel oxide (TO) formed by Rapid Thermal Oxidation, an $Si_3N_4$ charge trap layer (CTL) formed by low-pressure chemical vapor deposition (LPCVD), an $Al_2O_3$ blocking oxide (BO) formed by chemical vapor deposition, and an n+ poly-Silicon gate. The device fabrication is discussed in detail in [11].

The non-ideal program time of CTF due to N dependency and $T_{gap}$ dependency is described in [8] affects weight update of CTF in the network. In brief, the experimental observations are:

(a) N Dependency: The $V_T$ reduces as $N$ increases (N = 1, 25, 50, 100, 500, 1000, 2000, 5000, and 10000) with a corresponding decrease in the pulse width (pw = 2.5 ms, 0.1 ms, 50 $\mu s$, 25 $\mu s$, 5 $\mu s$, 2.5 $\mu s$, 1.25 $\mu s$, 0.5 $\mu s$, and 0.25 $\mu s$) The total ON time is conserved as 2.5 ms. The $V_T$ abruptly vanishes for smaller pulse (N > 1000 or pw > 2.5 $\mu s$) widths. Therefore, N cannot be further future divided into smaller $pw$, and there is a limit to N or pw.

(b) $T_{gap}$ Dependency: There is a strong logarithmic $t_{gap}$ dependence of $V_T$. When $t_{gap}$ is small (<$t_{critical}$), the $V_T$ increases as pulses are interactive due to charge accumulation in blocking oxide, For $t_{gap}$ longer than a critical time ($t_{critical}$), the $V_T$ saturates i.e., $V_T$ is independent *of* $t_{gap}$. Thus, a longer $t_{gap}$ resolves the interaction between the pulses.

## III. IMPACT OF NON-IDEAL PROGRAM TIME OF CTF

The three main impacts of non-ideal program time when employing CTF as an analog memory unit in RPU architecture are described in [8]:

1. The stochastic encoding error is reduced with the increase in $N$ ($\Delta_{stoch}$) as $\sigma/\mu = 1/\sqrt{N}$ [12]. However, the non-ideal program time of CTF has N-dependency, $V_T$ vanishes abruptly after a particular N (> 1000), which creates a limitation on $N$, and after that $V_T$ becomes ineffective.

2. The N dependency of non-ideal program time determines a systematic logarithmic increasing error in $V_T$. However, the $t_{gap}$ is randomly changing for every bit stream for stochastic multiplication in RPU. Hence, it adds stochastic noise to the final weight update.

These systematic and random errors due to CTF need to be compensated [8].

3. The Resistive Processing Unit (RPU) and other in-memory computing training systems frequently employ the division of an input pulse into an N-slot pulse train to stimulate the stochastic gradient descent, i.e., coarse updates initially followed by finer weight updates [13]. Therefore, it is crucial to model the $V_T$ which includes the N and $T_{gap}$ dependencies, and study its impact on neural networks.

To test the impact of non-ideal program time on stochastic variance, we used the experimental setup described in [8]. In brief, all input pulses biasing are applied to the gate terminal with respect to the substrate of the CTF device. The initial state of the device ($V_{T,0}$) has been read by using (-2, 2) V sweep before giving the pulse(s). A $T_{ON} = 2.5\ ms$ program pulse of magnitude $12.5\ V$ is applied and the $V_T$ is measured again using a $[-2,2]\ V$ sweep, and $10\ KHz$ capacitance-voltage measurement using the constant capacitance, $C_{FB} = 15\ pF$, method. The total ON time can be divided into $N$ discrete pulses of the same magnitude and pulse width, $t_{pw}$, such that $T_{ON} = N \times t_{pw}$. This experiment of initialization and programming is performed for different proposed pulse schemes mentioned in Table I.

TABLE I
COMPENSATION BY PULSE DESIGN

| Pulse Design (3 Consecutive Pulses) | $T_{gap}$ | $V_T$ Shift Error | Pulse Interaction | Computing Time | Design Cost |
|---|---|---|---|---|---|
| P1 No Gap, No Trap time | No (True) | Not resolved | Not resolved | Low | Low |
| P2 | > $T_{critical}$ (sparse) | Not resolved | Resolved | High | Low |
| P3 $t_{trap}$ | > $T_{critical}$ (sparse) | Resolved | Resolved | High | Low |
| P4 $t_{trap}$, $P_{de-trap}$ | No (True) | Resolved | Resolved | Moderate | High |

## IV. PULSING SCHEME TO COMPENSATE UNCONSERVED NON-IDEAL PROGRAM TIME

The total error in the weight update is the addition of the $V_T$ error and the stochastic error. The stochastic error ($\Delta_{stoch}$) as $\sigma/\mu = 1/\sqrt{N}$ reduces with the increase in $N$, whereas on the contrary, the $V_T$ error due to non-ideal program time increases with an increase in N [8]. The addition of these two errors makes the total error independent of N after a particular N (>30), causing an error in flooring (Noise Floor) as shown in Fig. 2 (red color). The $V_T$ error is due to charge accumulation in traps of BO of CTF (orange color). When these trap charges are not removed before appearing of the next pulse, these BO traps add on the final $V_T$, hence the pulses are interactive due to these unremoved trapped charges in the blocking oxide of CTF. Therefore, to resolve the $V_T$ error a trapping time ($t_{trap}$) must be added to the pulses. To avoid the pulse interaction de-trapping time ($t_{de-trap}$) must be provided as the gap between the pulses. It is the de-trapping time of the charge accumulated in the blocking oxide of CTF and is ≥ $t_{critical}$ for a particular pulse width. Hence, we propose compensation by pulse designs as shown in Table 1 to improve the error floor, These designed pulses can be used as the input pulses for CTF. The design of the pulses must compensate for the error floor shown in Fig.2. The simple pulse used for experiments in [8] is shown in Table 1 as pulse P1 (true $t_{gap}$) which produces $V_T$ error and interaction of pulses. In the second pulse scheme P2 (sparse $t_{gap}$), the gap between the pulses is ≥ $T_{critical}$ to avoid the interaction of pulses at the cost of computing time which is increased by ≥ $N \times$ Tcritical. As $V_T$ reduction is unresolved in P2, we propose pulse design P3, based on a 2-level compensation scheme.

(i) To mitigate the pw or $N$ pulse effect, $t_{trap}$ needs to be added to compensate for the $V_T$ error due to charge accumulation in traps of BO as mentioned in [8].

(ii) Further, to compensate for the interaction of pulses due to $t_{gap}$ variation, $t_{gap} \geq t_{critical} = t_{de-trap}$ [8] must be stipulated to reduce $t_{gap}$ variation at the cost of additional computing time.

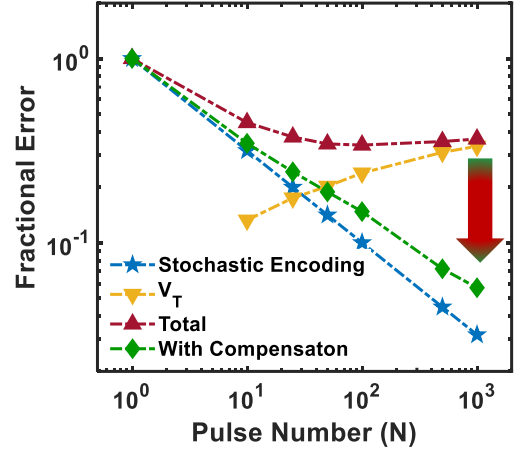

Fig. 2. Fractional error: Fractional error due to the stochastic encoding in pulse trains of length $N$ (blue) and error in $V_T$ is due to pulse division (orange). The addition of both errors is the total error for a network (red color), resulting in a noise floor. After doing the proposed compensation, there is an improvement in the noise floor of the total fractional error (green color) shown by the arrow.

The $t_{trap}$ (shaded area of P3, P4) and $t_{de-trap}$ (the gap between the pulses) for P3 is estimated by (1) and (2) respectively. The detailed method for estimation is described in [8].

$$t_{trap} = t_{pw} - T_{NV}/n_{req} \quad (1)$$
$$t_{de-trap} = t_{critical}(pw) \quad (2)$$

where $t_{pw}$ is the width of the pulse; $T_{NV}$ is an effective non-volatile writing time, $n_{req}$ is the number of pulses of width $t_{pw}$ needed to achieve a particular $V_T$. We have used $t_{de-trap} = 10\ ms$ which is taken as the maximum of all $t_{critical}$ for all *pws* used, as described in [8]. Thus, the $V_T$ error and pulse interactions are resolved by the P3 scheme, at the cost of computing time.

To reduce the cost of computing time, the fourth pulse scheme P4 is proposed. In pulse P4, we add the trap time $t_{trap}$ (shaded) to mitigate the $V_T$ reduction, and a negative de-trapping pulse ($P_{de-trap}$) is added with every pulse to avoid the interaction of pulses. The negative pulse pulls out the stored charges from blocking the oxide of CTF.

### A. Benefits of Compensation

We have used the pulse scheme P3 from Table I for compensation, the $t_{trap}$ is pw dependent, estimated by experimental data mentioned in [8] as (1) and $t_{gap}$ as $t_{de-trap}$ as (2). The addition of trap time and de-trap time in the pulse reduces the total error, thus improving the noise floor at the cost of computing time. The total computing time of the pulse train increases by $N \times (t_{trap} + t_{de-trap})$ due to the addition of $t_{trap}$ and $t_{de-trap}$ times in each pulse. Due to the experimental limitation for negative pulse generation as required in P4, we did not test the P4 pulse scheme, and is open for future experimental testing.

## V. Effect of Non-Ideal Program Time of CTF on Stochastic Matrix Multiplication

While error and its partial compensation are presented from a device perspective, the impact on systems performance of errors is required to find the device performance specifications for system-level operation. Hence, to study the impact of the non-ideal program time on weight update, $V_T$ of CTF in RPU, the CTF's $V_T$ is first mathematically modeled as weight update function to incorporate the effect of pulse number (N) or width (pw) and the gap between the pulses ($t_{gap}$). Later, we use this mathematical model of CTF's weight update for the stochastic matrix multiplication in RPU.

### A. Mathematical Modelling of Weight Update ($dV_T/dn$)

Mathematical modeling is done by curve-fitting the experimental data using appropriate equations, as shown in Fig 3. The curve-fitting for $V_T$ is done by separating the variables ($N, pw$ and $t_{gap}, pw$) and then modeling them separately as (3). First, an ideal weight-update function ($P$) is modeled as (4), which only incorporates the effect of N, pw on $V_T$ as shown in Fig. 3 (a) for linear scale, and (b) for log scale as (4). Next, a multiplier function ($Q$) is modeled, which incorporates the effect of the gap between the pulses ($t_{gap}$) as (5) as shown in Fig.3c. The separation of variables is only possible because $t_{gap}$ and $V_T$ are independent of each other. Hence the weight update function per pulse (n) is given by:

$$\frac{dV_T}{dn} = P(V_T, pw) * Q(t_{gap}, pw) \quad (3)$$

$$P(V_T, pw) = A \exp(BV_T) \quad (4)$$

$$Q(t_{gap}, pw) = \begin{cases} C_1 \log(t_{gap}) + C_2 & t_{gap} < T_{critical} \\ C_1 \log(T_{critical}) + C_2 & t_{gap} > T_{critical} \end{cases} \quad (5)$$

where, $A, B, C_1$ & $C_2$ are curve-fitting constants that strongly depend on pw. They are also subject to device process, voltage, and Temperature (PVT) variations. In earlier work [6], the effect of $t_{gap}$ was not considered and the Q function was set to a constant value. $T_{critical}$ used in the Q function corresponds to the point beyond which variance in the effect of $t_{gap}$ is not observed. It is termed as a sparse $t_{gap}$ region as described in pulse scheme P2 and P3 of Table I. Q function shows the highest correlation in the piecewise linear model.

The given model has been fit into 4 different sets of experimental data corresponding to different pw to verify the modeling function. Further, the 4 cases of N (1000, 500, 100, and 25) and pw (2.5, 5, 25 and 100)$\mu s$ respectively such that total $T_{ON}$ time is always equal to 2.5 ms [8] have been compared to observe variations among the same for stochastic computing.

### B. Stochastic Matrix Multiplication for Computing

Weight update in an RPU takes place in the form of stochastic pulse trains, where the number of ON pulses in the train is proportional to the weight update. The update value for a weight, $W_{ij}$ connecting neuron $i$ in the pre-layer to neuron $j$ in the post-layer is calculated by stochastic multiplication of the pre-layer activation $x_i$ and post-layer gradient $\delta_j$ where both are represented by independent stochastic pulse trains [4]. This implements Backpropagation with a $1/\sqrt{N}$ dependent stochastic gradient error [12], where $N$ is the length of the stochastic pulse train.

Given the pulse train and the initial $V_T$, the final $V_T$ is calculated using the algorithm given in Pseudocode 1. It iteratively updates the weight after each pulse by calculating the $dV_T$ based on current $V_T$ and $t_{gap}$. Note: Two consecutive pulses are separated by a $t_{gap}$ of $\epsilon$, which is infinitesimally small. This helps in the solution of the modeled equation.

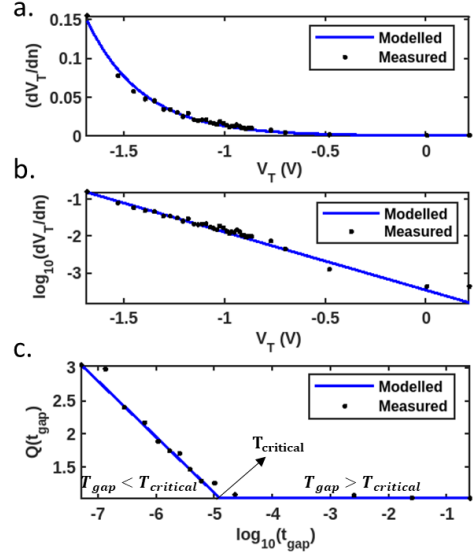

Fig. 3. Curve fitting on experimental data of P function with sparse $t_{gap}$ and pw = 2.5$\mu s$ in (a) linear, and (b) logarithmic scale shows that the effect of $V_T$ is exponential. (c) Curve fitting of multiplier function Q shows the effect of $t_{gap}$ to be piecewise linear.

TABLE II
PSEUDOCODE 1: STOCHASTIC COMPUTING WEIGHT UPDATE:

| Algorithm: $V_T$ update corresponding to a Stochastic Pulse Train |
|---|
| **Input**: Stochastic Pulse Train $x^{(i)}$ ; Weight update function $dV_T(V_T, t_{gap})$; Initial Threshold Voltage ($V_{T0}$) ; Pulse Width ($pw$) |
| **Output:** Final Threshold Voltage ($V_{TF}$) |
| Let $V_T = V_{TO}$ be the device threshold voltage |
| $\epsilon$ = infinitesimally small unit of time |
| **For each** ON pulse ($x^{(i)} = 1$) **do** |
|     Let $\Delta x$ be the distance from the last ON Pulse |
|     $t_{gap} = \epsilon + \Delta x * pw$ |
|     $V_T \leftarrow V_T + dV_T(V_T, t_{gap})$ |
| **End** |

The given algorithm was run a sufficient number of times from initial $V_{T0} = 0V$ for different weight updates by stochastic computing with probabilities of $x^{(i)} = [0, 0.2, 0.4, 0.8, 1]$, for above mention 4 cases of N [1000, 500, 100, 25], by dividing the pulse width (pw) into pw, 2.5, 5, 25 and 100 $\mu s$ respectively such that the total on time [$T_{ON} = 2.5ms$] is fixed. The mean and variance of the weight update model were obtained for three cases based on $t_{gap}$, namely, ideal, sparse, and true. Note: $N$ and pw effects can be used interchangeably as they have a fixed product ($= T_{ON}$) [8].

These three cases are shown in Fig. 4. In case 1, $V_T$ update is purely decided by the weight update, $T_{gap}$ and $N$ effects are ignored. This has a non-linearity due to $V_T$ dependence on weight update. In case 2, the $N$ effect is added and the $T_{gap}$ is set to large (sparse). The $V_T$ degrades with an increase in $N$, owing to a higher number of BO trapping [7]. In case 3, the $T_{gap}$ effect is captured. The $T_{gap}$ changes between every pair of pulses and in every run due to the stochastic nature of the pulse train. However, the $T_{gap}$ effect (case 3) improves the linearity and range as shown in Fig.4a whereas for sparse $t_{gap}$ (case 2) degradation of $V_T$ varies with N. The variability ($\sigma/\mu$) of stochastic computing shown in Fig 4b is lower for higher N and higher for true $t_{gap}$.

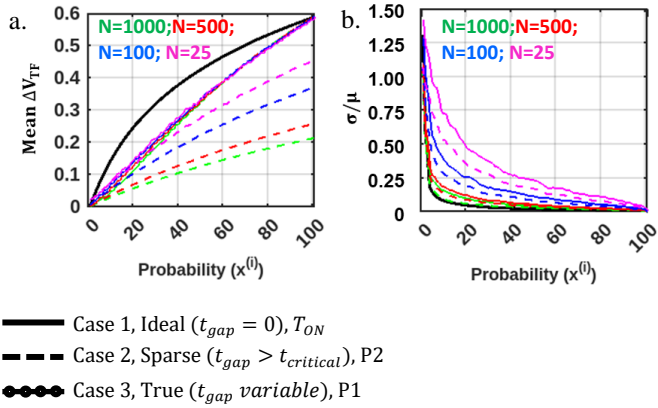

— Case 1, Ideal ($t_{gap} = 0$), $T_{ON}$
--- Case 2, Sparse ($t_{gap} > t_{critical}$), P2
●●●● Case 3, True ($t_{gap}$ variable), P1

Fig. 4. The change in threshold voltage $\Delta V_T$ vs $x^{(i)}$ probability for (i) Ideal (ii) only $PW$ dependent (iii) $PW$ and $t_{gap}$ dependent (realistic) – cases is observed for its (a) mean value and (b) $\sigma/\mu$ variation for different N.

## VI. RPU-BASED NEURAL NETWORK TRAINING USING CTF

To test the performance of RPU-based neural network training with the non-idealities of CTF, we trained the fully connected neural network shown in Fig.1d with 784 input, 1280 hidden, and 10 output neurons for the supervised classification of digits in the MNIST dataset and fashion MINST data set.

### A. MNIST

To test the performance of the neural network with the non-idealities of CTF, we trained the neural networks for the supervised classification of digits in the MNIST dataset [9]. The MNIST dataset consists of $60000$ training and $10,000$ test images of $10$ handwritten digits, each of size $28 \times 28$ pixels. A fully connected neural network with 1 hidden layer consisting of $1280$ neurons, was used for classification. Weights in the network are represented by ($W_p - W_n$) where both $W_p$ and $W_n$ contain only positive values. Weights were constrained to move along the mean $\Delta V_T$ trajectory shown in Fig. 4(a), scaled and mapped to $[0, 1]$. This is referred to as the "**non-linearity**" of the weight updates. The raw weight updates $\Delta W$ were quantized by the value of $N$ and corrupted with Gaussian **noise** as characterized in Fig. 4(b), and then cast in the $V_T$ trajectory function. Positive weight updates ($\Delta W > 0$) were applied to $W_p$ and the absolute values of the negative weight updates ($|\Delta W < 0|$) were applied to $W_n$. The effect on learning performance for different N and $T_{gap}$ configurations is shown in Fig 5.

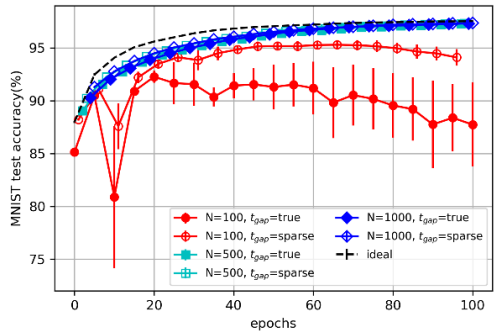

Fig. 5. Performance of RPU-based neural network training to recognize the MINST data with CTF weight non-idealities for different N (100, 500, and 1000) and with true $T_{gap}$, and sparse $T_{gap} > 1\mu s$.

### B. Fashion-MNIST

Fashion-MNIST [10] is another image classification dataset. Like MNIST, it also consists of $60000$ training and $10,000$ test images of $10$ types of clothing items, each of size $28 \times 28$ pixels. The rest of the simulation details are identical to what has been described above for the MNIST dataset. The effect on learning performance for different values of N with test accuracy is shown in Fig. 6

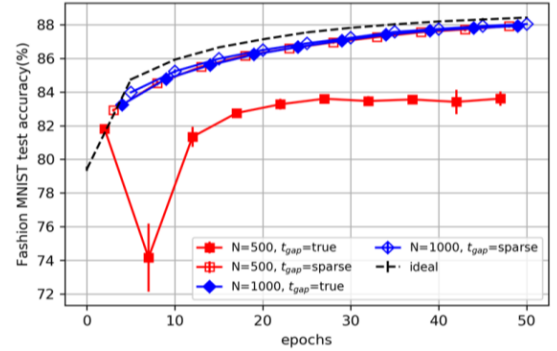

Fig. 6. Performance of RPU-based neural network training to recognize the Fashion MINST data with CTF weight non-idealities for different N (500 and 1000) and with true $T_{gap}$, and sparse $T_{gap} > 1\mu s$.

### C. Observation of Training Performance and Noise Analysis

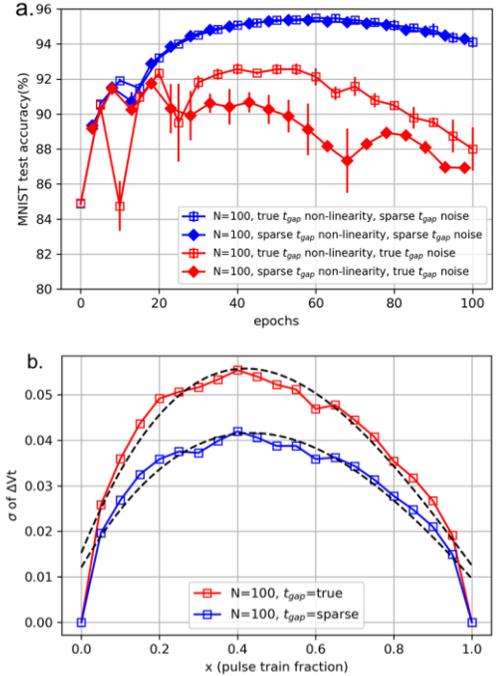

Fig. 7. (a) Network training performance for N=100, with all combinations of noise and non-linearity for true $T_{gap}$ and sparse $T_{gap}$ cases. We find that high noise simulations (true $T_{gap}$ noise level) are significantly worse than low noise simulations (sparse $t_{gap}$ noise level) irrespective of the kind of non-linearity. (b) The noise level ($\sigma$ of $\Delta V_T$) is lower for sparse $t_{gap}$ compared to the true $t_{gap}$ for lower N (=100).

We find that the test accuracy for higher $N$ (~1000) is unaffected by the $T_{gap}$, whereas for lower $N$, i.e., 100 for MNIST and 500 for Fashion-MNIST, performance degrades for true $T_{gap}$ as opposed to sparse $T_{gap}$ as shown in Fig. 5 and Fig. 6 respectively. To analyze the network training performance at lower N, an ablation study is performed by simulating RPU-based neural network training with all the possible combinations of "non-linearity" and "noise" CTF for lower N (=100). We find that neural network training simulations with low noise (sparse $t_{gap}$), CTF have better performance as shown in Fig.

7(a), where the difference caused by the different non-linearities is negligible. In contrast, in neural network training simulations with high noise (true $T_{gap}$) shown in Fg.7(b). CTF characteristics have worse performance and observable differences arising from the different non-linearities as shown in Fig. 7(a). In a nutshell, we can conclude that the performance of the network can be improved for lower N, by using the sparse $t_{gap}$ noise and independent of non-linearity as shown in Fig. 7(a). This is due to the lower noise of $\sigma$ of $V_T$ for sparse $t_{gap}$.

TABLE II
OVERALL COMPARATIVE ANALYSIS OF NEURAL NETWORK PERFORMANCE

| N | Metric | MNIST | | FMNIST | |
|---|---|---|---|---|---|
| | | True $t_{gap}$ | Sparse $t_{gap}$ | True $t_{gap}$ | Sparse $t_{gap}$ |
| 100 | Latency (per sample) (ms) | 2.5 | 1002.5 | Not Converge | |
| | Energy (per sample) (J) | $5.96 \times 10^{-14}$ | $6.39 \times 10^{-14}$ | | |
| | Accuracy (%) | $88.5 \pm 1.4$ | $93.5 \pm 0.3$ | | |
| | Sigma of $\Delta V_T$ | 0.06 | 0.04 | | |
| 500 | Latency (per sample) (ms) | 2.5 | 5002.5 | 2.5 | 5002.5 |
| | Energy (per sample) (J) | $5.70 \times 10^{-14}$ | $5.52 \times 10^{-14}$ | $1.38 \times 10^{-14}$ | $1.34 \times 10^{-14}$ |
| | Accuracy (%) | $97.31 \pm 0.03$ | $97.34 \pm 0.01$ | $83.4 \pm 0.9$ | $87.57 \pm 0.01$ |
| | Sigma of $\Delta V_T$ | 0.025 | 0.018 | 0.025 | 0.018 |
| 1000 | Latency (per sample) (ms) | 2.5 | 10002.5 | 2.5 | 10002.5 |
| | Energy (per sample) (J) | $5.80 \times 10^{-14}$ | $5.49 \times 10^{-14}$ | $1.36 \times 10^{-14}$ | $1.29 \times 10^{-14}$ |
| | Accuracy (%) | $97.35 \pm 0.04$ | $97.37 \pm 0.01$ | $87.4 \pm 0.17$ | $87.77 \pm 0.06$ |
| | Sigma of $\Delta V_T$ | 0.02 | 0.012 | 0.02 | 0.012 |
| Ideal | Accuracy (%) | $97.51 \pm 0.03$ | | $88.1 \pm 0.07$ | |

D. Energy and Latency Performance of Neural Network

Energy: The energy requirement for the weight update operation on a scaled 180 nm CTF device is estimated by counting the average number of update pulses that are applied to the weight matrices in the network for any given training sample and multiplying that by the energy required for each update pulse. The results of that energy estimation are shown in Table II. We observe that the no significant difference between the energy requirements for the *true* and *sparse* $T_{gap}$ case. We observe that in comparison to the state-of-the-art described in [14], the energy requirement of CTF (70 fJ) is comparable to the human brain energy of 10 fJ and significantly (20k times) less while using CTF as an AMU in RPU compared to analog ReRAM for the outer product (2 nJ) [14].

Latency: The latency of the network is significantly higher in the *sparse* $T_{gap}$ case when compared with the *true* $T_{gap}$ ($10^3$ times) case for both MNIST and FMNIST data sets as shown in Table II. This is because of the addition of $t_{de-trap}$ time for each pulse in the case of *sparse* $T_{gap}$ case. However, it is higher compared to the state-of-the-art analog ReRAM-based learning [14], although comparable to the human brain latency in the range of ms [15].

VII. CONCLUSIONS

In this work, the impact of the experimentally observed non-ideal program time in CTF is analyzed, characterized, and its effects on the training of deep neural networks are studied. The noise floor in fractional error is caused by variance of stochastic computation and error due to non-ideal program time is experimentally compensated by adding the trap and de-trap time in each input pulse. We model the characteristics of non-ideal program time and use it for stochastic computing. The weight update from stochastic computing is used to train the DNN with RPU architecture for MNIST and fashion MNIST datasets. The training performance is best for a larger N, and independent of $t_{gap}$, as specified in the state-of-art for RPU element, $N$ must be greater or equal to 1000. However, in our study we showed that the accuracy can be improved even for lower ($N < 500$), by keeping sparse $t_{gap}$ ($> t_{critical}$). This study shows that even for lower N, we can achieve good learning accuracy by careful pulse design such that $t_{gap}$ is ensured to be greater than $T_{critical}$ at the cost of higher latency. The energy requirement by CTF as AMU in RPU architecture is 70 fJ, comparable to the human brain energy of 10 fJ and significantly lesser to other state of art like analog ReRAM. Such a study enhances the system-level feasibility of using input pulses for CTF-based RPU architecture toward large-scale ANN with accurate on-chip training for in-memory computing.